\documentclass[times, twoside]{zHenriquesLab-StyleBioRxiv}

\leadauthor{Patton, Zouinkhi, \& Rhoades} 

\begin{document}

\title{DaCapo: a modular deep learning framework for scalable 3D image segmentation}
\shorttitle{DaCapo}


\author[1,$\dagger$]{William Patton}

\author[1,$\dagger$]{Jeff L. Rhoades}

\author[1,$\dagger$]{Marwan {Zouinkhi}}

\author[1]{David G. Ackerman}

\author[1]{Caroline Malin-Mayor}

\author[1]{Diane Adjavon}

\author[1]{Larissa Heinrich}

\author[1]{Davis Bennett}

\author[1]{Yurii Zubov}

\author[1]{CellMap Project Team}

\author[1,\href{mailto:weigela@janelia.hhmi.org}{\Letter}]{Aubrey V. Weigel}

\author[1,\href{mailto:funkej@janelia.hhmi.org}{\Letter}]{Jan Funke}

\affil[1]{Janelia Research Campus, HHMI, 19700 Helix Drive, Ashburn, 20147, VA, USA}
\affil[$\dagger$]{These authors contributed equally.}
\affil[ \href{mailto:weigela@janelia.hhmi.org,funkej@janelia.hhmi.org}{\Letter}]{corresponding authors}

\maketitle

\begin{abstract}
DaCapo is a specialized deep learning library tailored to expedite the training and application of existing machine learning approaches on large near-isotropic image data. In this correspondence, we introduce DaCapo's unique features optimized for this specific domain, highlighting its modular structure, efficient experiment management tools, and scalable deployment capabilities. We discuss its potential to improve access to large-scale, isotropic image segmentation and invite the community to explore and contribute to this open-source initiative.
\end{abstract}

\begin{keywords}
deep learning | machine learning | segmentation | big data | open source
\end{keywords}

\begin{corrauthor}

\texttt{\href{mailto:weigela@janelia.hhmi.org}{weigela@janelia.hhmi.org}}, \texttt{\href{mailto:funkej@janelia.hhmi.org}{funkej@janelia.hhmi.org}}
\end{corrauthor}

\section*{Introduction}
Extracting meaningful biological insights from imaging data is greatly enhanced by the accurate segmentation of structures such as cells and organelles. With the continuous advancements in imaging technologies and techniques, the increasing size, dimensionality, and information density of collected images pose significant challenges to scaling existing machine learning approaches for image segmentation. This is especially true for volume electron microscopy imaging methods, especially those with near-isotropic capabilities such as focused ion beam - scanning electron microscopy (FIB-SEM). Conventional 2D neural network-based segmentation approaches are not fully optimized for these imaging modalities.

To address these challenges and meet the demands for both scalability and 3D-aware segmentation networks, we developed DaCapo, a modular and open-source framework designed for training and deploying deep learning solutions at scale. DaCapo efficiently handles terabyte- and teravoxel-sized datasets by integrating established segmentation methodologies with blockwise distributed deployment across local, cluster, or cloud infrastructures.

\begin{figure*}
\centering
\includegraphics[width=0.95\textwidth]{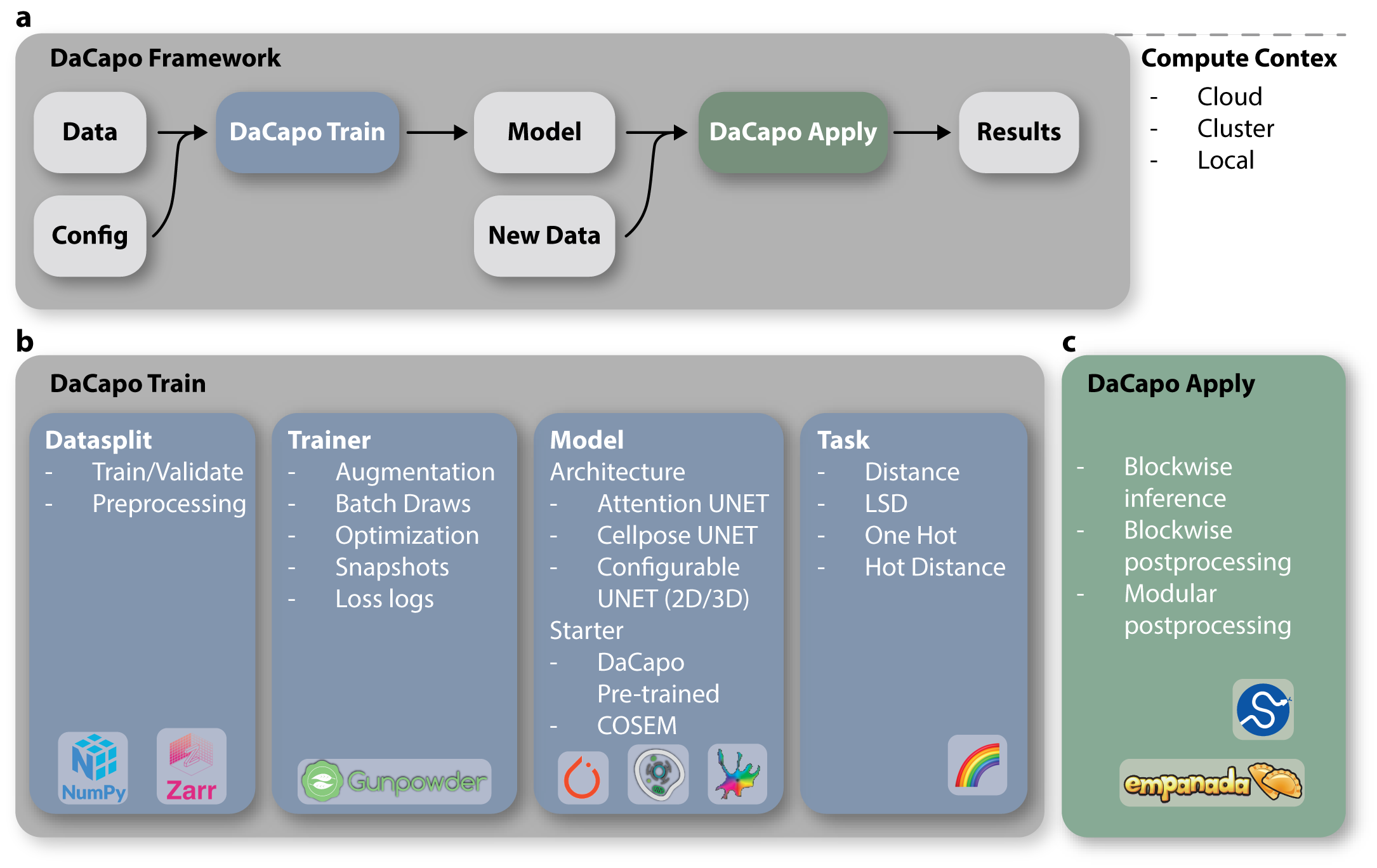}
\caption{\textbf{Anatomy of DaCapo.} \textbf{\textit{a) DaCapo framework}} is subdivided into individually configurable submodules. For model training, DaCapo takes information about data sources and configuration of training hyperparameters, such as the number of iterations or batch size (see panel b for more details). Trained models can then be quickly applied to new datasets. Both training and post-training tasks can be run in various compute contexts, utilizing the power of cloud and cluster resources, as well as the simplicity of local setups. \textbf{\textit{b) DaCapo train}} requires specification of a datasplit, trainer, model, and task. The dataplit specifies which datasources should be used for training vs. validation scores. This can be supplied with a simple CSV file. Training hyperparameters such as number of training iterations, data augmentations, etc. are also easily configurable in the trainer. The network architecture to be used (model) and target representation the model is being trained to predict (task) are also specified here. \textit{\textbf{c) DaCapo Apply}} is used after training is complete. Prediction and post-processing can be seamlessly scaled with DaCapo via blockwise processing of any size volume.}
\label{fig1}
\end{figure*}

DaCapo's functionality is highly adaptable, with submodules that can be tailored to the user's specific requirements. This includes options for 2D or 3D segmentation, handling both isotropic and anisotropic data, and selecting between tasks such as semantic or instance segmentation. Users can also choose their preferred neural network architectures, including pretrained models, apply various data augmentations (e.g., rotations), and decide on the compute infrastructure (local, cluster, or cloud), blockwise batch processing methods (inference or post-processing), and data storage solutions (file system or databases). See Figure \ref{fig1}.

In the subsequent sections, we demonstrate the application of DaCapo, particularly focusing on its utility for FIB-SEM data, showcasing its versatility and effectiveness in managing and segmenting complex imaging datasets.

\section*{Training Setup}
DaCapo manages the process of training, model checkpointing, and post-processing parameter selection. Using a simple tabulated entry (e.g. CSV) users can designate subsets of their data to be used for training or validation. Data loading, image augmentation, loss calculation, and model parameter optimization are all performed under the hood by DaCapo, utilizing Gunpowder\cite{noauthor_undated-wz}. Simultaneously, DaCapo periodically gathers validation scores for the current model on held out data, performing simple parameter sweeps for the post-processing approach used (e.g. varying the cutoff value for thresholding). Many metrics for evaluating validation performance are included, such as F1-score and Jaccard index for semantic segmentation, and Variation of Information (VoI) for instance segmentation (see DaCapo documentation for the complete list: \url{dacapo.readthedocs.io}). Validation scores, loss scores, and processed predictions are stored for use during model selection. The optimal iteration and best performing parameters (e.g. threshold) are stored for ease of reference and prompt implementation.

\section*{Task Specification}
With a simple change to a single line of code, DaCapo can switch from semantic segmentation setup, to one-hot encoding predictions, to predicting signed tanh boundary distances of the binary labels\cite{Heinrich2018-bb, Heinrich2021-ap} or to using the hot-distance approach\cite{Zouinkhi2024-sm}, which combines one-hot and signed boundary distance embeddings for segmentation. In addition to these prediction targets designed for semantic segmentation, DaCapo also features targets designed for instance segmentation, including prediction of local shape descriptors\cite{Sheridan2023-nh}, as well as long and short-range pixel-wise affinities\cite{Lee2017-ar, Sheridan2023-nh}. Furthermore, DaCapo's modularity has been specifically designed to allow for easy addition of new prediction targets to its arsenal, so as to maintain state of the art functionality.

\section*{Model Architecture}
Multiple model architectures are pre-built into DaCapo, including 2D and 3D UNet variants, such as the Cellpose model \cite{Pachitariu2022-vy,Stringer2021-fa}. DaCapo is also designed to use models previously trained by the user or take advantage of pretrained models available for download. Included, are a number of existing neural networks trained for segmentation of cells and subcellular structures in FIB-SEM images by the COSEM Project Team\cite{Heinrich2021-ap} (see \href{https://openorganelle.org}{openorganelle.org}). These can be downloaded as starter models and further finetune, \ref{fig1}b. More information can be found in the \href{https://janelia-cellmap.github.io/dacapo/autoapi/dacapo/experiments/starts/index.html}{starter documentation}. These pretrained 3D UNets have demonstrated utility as general purpose feature extractors that can be finetuned on a number of other FIB-SEM datasets. Future CellMap models will continue to be included as optional downloads within DaCapo. We also welcome and encourage other members of the community to contribute their state of the art models to DaCapo's repertoire.

\section*{Blockwise Inference \& Post-processing}
In order to scale deployment to petabyte-scale datasets, DaCapo employs blockwise inference and post-processing steps using Daisy\cite{Nguyen2022-eo}. This allows it to handle datasets that cannot fit into memory all at once, while also eliminating edge artifacts. By leveraging chunked file formats (i.e. Zarr-V2\cite{Miles2020-pl} and N5\cite{noauthor_undated-kj}), DaCapo is able to seamlessly parallelize both semantic and instance segmentation. For instance segmentation, this includes a secondary relabeling step that unifies object IDs across all data chunks, so that no additional merging mechanisms are necessary after DaCapo's post-processing is complete.

Blockwise processing has also been implemented in a generalizable manner, allowing users to write simple custom scripts for tailored post-processing solutions and apply them to big data. These scripts can operate as simply as taking in a small Numpy-style\cite{Harris2020-or} array of values to be processed, and outputting another small array with the processed results. In practice, this means users do not need expertise with chunked file formats or parallelization to scale clever post-processing solutions. As an example, we have implemented blockwise application of Empanada\cite{Conrad2023-yx} to segment mitochondria in arbitrarily large image volumes. We hope to see additional community solutions scaled with the help of DaCapo in the future.

\section*{Compute Contexts}
DaCapo features globally implemented compute context configuration. This allows easy specification of factors such as whether operation should be handled locally on a single node or distributed to a compute cluster, and the number of CPUs and GPUs to use during training, inference and blockwise processing. Data and results can be stored locally or with a number of cloud protocols (e.g. s3, gs, http, etc.)\cite{noauthor_undated-ni}, providing flexibility depending on the needs of the project. Currently, local GPU and CPU, as well as LSF-managed cluster instances are supported, and the addition of custom compute environments is easy because of DaCapo's design. Additionally, a Docker image is provided for easy deployment on AWS and similar cloud resources.

\section*{Conclusion}
We present DaCapo as a valuable asset for researchers and practitioners working with large volume image data, offering a dedicated platform for efficient model training and deployment. DaCapo is tailored for users looking for customizable and scalable solutions for biological image segmentation, with an emphasis on long-term data accumulation and model generalization.

Our immediate next steps include enhancing the platform's user interface for more intuitive interaction, expanding the repository of pretrained models, and optimizing the system for even greater scalability and efficiency. This includes removing the current limit on instance segmentation, which restricts users to a total of $2^{64}$ unique objects per class. We invite researchers and practitioners to follow our progress and contribute to the ongoing development of DaCapo by visiting our \href{https://github.com/janelia-cellmap/dacapo}{repository}. Stay updated with our latest developments and join our community in advancing biological image analysis.

\begin{acknowledgements}
We would like to express our gratitude to all the contributors to our GitHub repository. Their valuable input, insightful feedback, and collaborative spirit have significantly enhanced the quality of DaCapo. We appreciate the time and effort invested by each contributor in reviewing code, suggesting improvements, and sharing their expertise. This work would not have been possible without their collective contributions. This work was supported by Howard Hughes Medical Institute, Janelia Research Campus.
\end{acknowledgements}

\begin{contributions}
W.P. and J.F. conceived and initially implemented DaCapo. J.L.R. and M.Z. lead subsequent development and testing, including blockwise processing and documentation, as well as preparing this manuscript. D.G.A., C.M.M., D.A., L.H., D.B., Y.Z., and the rest of the CellMap Project Team contributed to general development and review of the project. A.V.W. and J.F. oversaw general project development. A.V.W., D.G.A., and D.B. gave meaningful contributions to the preparation of this manuscript. The CellMap Project Team during this time was: David Ackerman, Emma Avetissian, Davis Bennett, Marley Bryant, Hannah Nguyen, Grace Park, Alyson Petruncio, Alannah Post, Jacquelyn Price, Diana Ramirez, Jeff Rhoades, Rebecca Vorimo, Aubrey Weigel, Marwan Zouinkhi, Yurii Zubov. Misha Ahrens, Christopher Beck, Teng-Leong Chew, Daniel Feliciano, Jan Funke, Harald Hess, Wyatt Korff, Jennifer Lippincott-Schwartz, Zhe J. Liu, Kayvon Pedram, Stephan Preibisch, Stephan Saalfeld, Ronald Vale, and Aubrey Weigel were part of the CellMap Steering Committee.
\end{contributions}

\begin{code}
The codebase is available on the GitHub repo: \href{https://github.com/janelia-cellmap/dacapo}{github.com/janelia-cellmap/dacapo}
\end{code}

\begin{supplementary}
Installation instructions, tutorials, and extended documentation for DaCapo is available on \href{https://janelia-cellmap.github.io/dacapo/}{GitHub.io} and \href{https://dacapo.readthedocs.io/en/stable/}{ReadtheDocs.io}.
\end{supplementary}

\section*{Bibliography}
\bibliography{paperpile}

\captionsetup*{format=largeformat}


\end{document}